\documentclass{trbunofficial}
\usepackage{graphicx}
\usepackage{adjustbox}
\usepackage{xstring}
\usepackage{amsmath}
\usepackage{amsfonts}
\usepackage{geometry} 
\usepackage{booktabs}
\usepackage{multirow}
\usepackage[table]{xcolor}
\usepackage{soul}
\soulregister\cite7 
\soulregister{\ref}{1}
\usepackage[most]{tcolorbox}

\usepackage[hidelinks]{hyperref}

\AuthorHeaders{ }
\usepackage{titlecaps} 

\title{Positional-aware Spatio-Temporal Network for Large-Scale Traffic Prediction}

\author{%
  \textbf{Runfei Chen}\\
   Ph.D. Student\\
   Urban Mobility Institute\\
   Tongji University\\
   Shanghai, China, 200092\\
   Email: runfeichen@tongji.edu.cn
}

\TotalWords{5956}

\begin{document}
\maketitle
\section{Abstract}
Traffic flow forecasting has emerged as an indispensable mission for daily life, which is required to utilize the spatiotemporal relationship between each location within a time period under a graph structure to predict future flow.
However, the large travel demand for broader geographical areas and longer time spans requires models to distinguish each node clearly and possess a holistic view of the history, which has been paid less attention to in prior works.
Furthermore, increasing sizes of data hinder the deployment of most models in real application environments.
To this end, in this paper, we propose a lightweight Positional-aware Spatio-Temporal Network (PASTN) to effectively capture both temporal and spatial complexities in an end-to-end manner.
PASTN introduces positional-aware embeddings to separate each node's representation, while also utilizing a temporal attention module to improve the long-range perception of current models.
Extensive experiments verify the effectiveness and efficiency of PASTN across datasets of various scales (county, megalopolis and state).
Further analysis demonstrates the efficacy of newly introduced modules either.

\hfill\break%
\noindent\textit{Keywords}: Traffic Prediction, Spatio-temporal network, Position embedding, Large-scale traffic
\newpage

\section{1 Introduction}
As vehicle numbers increase, road congestion emerges as a significant barrier to societal efficiency~\cite{wang2020control,xiao2021vehicle}, enhancing greenhouse gas emissions and elevating accident risks. Accurate traffic prediction is essential to improve traffic management and increase operational efficiency. Such forecasting supports crucial services including intelligent route planning and dynamic traffic management, which in turn mitigate the adverse effects of congestion~\cite{wu2016short}. With daily commutes and travel extending across local borders and affecting counties, cities, and entire states~\cite{clark2016changes,schlapfer2021universal}, forecasting traffic flow across broader regions becomes indispensable.

The challenges of traffic prediction mainly stem from the complex spatiotemporal correlations between flow sensors~\cite{yin2021deep}. In recent years, deep neural networks have emerged as the preferred method for traffic forecasting. The use of Graph Neural Networks (GNNs) to model spatial dependencies, combined with Recurrent Neural Networks (RNNs), Temporal Convolutional Networks (TCNs) for temporal analysis, has become increasingly prevalent~\cite{li2023dynamic}. 
Such frameworks extract complex spatio-temporal dependencies of both local adjacency and broader, remote correlations across different times and regions. For example, traffic flow is not only influenced by local temporal patterns but also by similar patterns in distant functional areas. Moreover, longer observation periods unveil distinctive seasonal features, such as the effects of climatic similarities or recurrent traffic events~\cite{su2023roformer,shao2022decoupled}. Enhancing model capabilities to interpret these comprehensive dynamics can significantly improve the utilization of traffic flow data.


While these developments have yielded promising results in forecasting traffic flows within small-scale, simple subdivision PEMS datasets~\cite{yu2018spatio,li2018diffusion,song2020spatial}, they have shown limited effectiveness on large-scale networks~\cite{liu2023largest, mallick2020graph}, which is increasingly important due to the large travel demand for traffic flow forecasting across larger geographical areas and longer time spans.
Unfortunately, large-scale traffic forecasting presents significant challenges yet unresolved. On one hand, managing individual subdivision networks within a large traffic system can lead to extensive computational and memory demands~\cite{mallick2020graph}. 
On the other hand, existing end-to-end models are not applicable on large datasets due to the quadratic complexity of spatial and temporal modeling~\cite{liu2023we}, which significantly escalates computational costs on datasets with a substantial number of nodes and time steps.
Furthermore, although widely adopted GNN-based models are capable of capturing large-scale spatial relationships, they tend to diminish node-specific characteristics when the modeling layers increase~\cite{rusch2023survey}, which leads to a suboptimal forecasting result. 

To address these issues, we introduce a lightweight Positional-aware Spatio-Temporal Network (PASTN) to capture both temporal and spatial complexities. 
PASTN builds upon foundational spatio-temporal learning modules, enhancing them with positional awareness. 
Spatially, we employ absolute positional embeddings to initialize encodings that learn unique node representations and mitigate the over-smoothing problems associated with GNNs. 
Temporally, an attention-based mechanism augments the TCN in the spatio-temporal learning module, improving the learning of adjacent time-step correlations and facilitating global time-step interactions, which further strengthen the modeling of characteristics of each data point. 
This approach allows for a comprehensive understanding of historical dynamics across extensive traffic data encompassing varied temporal spans. 

Overall, our contributions can be summarized as follows:
\begin{itemize}
    \item We introduce a novel method that is aware of both time and space. The spatial-aware embedding layer provides a simple yet effective solution that circumvents the complexities of modeling inter-node relationships in large-scale networks. Additionally, the use of attention-enhanced modeling instead of solely relying on TCNs allows for robust temporal feature extraction, achieving comprehensive modeling of both long-range and short-range temporal dependencies in large datasets.
    \item Extensive experiments demonstrate that PASTN significantly outperforms state-of-the-art (SOTA) methods, achieving up to an 18.45\%(in RMSE) improvement without a substantial increase in learnable parameters and inference time.
    \item We also explore model performance during important times (e.g., peak morning and evening hours) and in critical regions (e.g., city centers), as well as its capabilities in long-term forecasting, consistently showing superior performance.
\end{itemize}

\section{2 Related Work}
\textbf{Traffic Flow Predicting.}
Recent advances in traffic forecasting have predominantly leveraged deep neural networks, supplanting traditional statistical methods such as ARIMA~\cite{williams2003modeling} and SVM~\cite{li2018brief}. These networks typically integrate Graph Neural Networks (GNNs) with either Recurrent Neural Networks (RNNs) or Temporal Convolutional Networks (TCNs) to model spatiotemporal dependencies inherent in traffic data. Prominent examples include the DCRNN~\cite{li2018diffusion}, ST-MetaNet~\cite{pan2019urban} and AGCRN~\cite{bai2020adaptive}, which utilize RNN variants to capture temporal dynamics. To enhance training efficiency and exploit parallel computation capabilities, several studies have replaced RNNs with TCNs equipped with dilated causal convolutions, as seen in STGCN~\cite{yu2018spatio}, GWNET~\cite{wu2019graph}, and DMSTGCN~\cite{han2021dynamic}. Despite the effectiveness, TCNs often struggle with unstable traffic flow variations due to limited receptive fields. Attention mechanisms, such as those in ASTGCN~\cite{guo2019attention} and PDFormer~\cite{jiang2023pdformer}, have been increasingly employed to model long-term correlations. However, these architecture, while robust in global feature extraction, suffer from stability issues due to high data volume demands~\cite{ivanov2021data}. In capturing spatial dynamics, two major trends have emerged. The first involves coupling GNNs with neural ordinary differential equations to produce continuous layers for modeling long-range spatio-temporal dependencies, e.g., STGODE~\cite{fang2021spatial} and STGNCDE~\cite{choi2022graph}. The second trend involves building adjacency matrices at different time-steps to capture the dynamic correlations among nodes, e.g., DGCRN ~\cite{li2021dynamic}, DSTAGNN~\cite{lan2022dstagnn}, and $\text{D}^{2}\text{STGNN}$~\cite{shao2022decoupled}. While these models show promising results on subdivision-level PEMS datasets, they struggle with scalability due to large memory consumption and computational costs when applied to larger datasets~\cite{liu2023largest}. 

Beyond refining the core network architectures, a significant trend in recent research involves enriching the modeling process with more complex data sources and structural priors to better capture real-world traffic dynamics.  Some studies move beyond single-modality sensor data, constructing multi-view graphs that incorporate information from public transit systems, surrounding Points-of-Interest, and other metadata~\cite{zhang2023incorporating,wei2024multi,fang2025heterogeneous}. Other works focus on the heterogeneity of nodes within the graph, proposing hierarchical structures or methods to identify and assign greater importance to pivotal nodes, such as major interchanges, which have a disproportionate impact on the overall traffic flow~\cite{guo2021hierarchical, huo2023hierarchical, kong2024spatio}. Another direction involves shifting from direct prediction to reconstruction-based frameworks~\cite{wang2018traffic, bilotta2023high,lian2025semi}. For example, ~\citet{liu2023spatio} utilize a spatio-temporal autoencoder to learn a latent representation of intrinsic traffic patterns before projecting it into the future. While these approaches introduce valuable, fine-grained perspectives, they often add extra layers of complexity. Consequently, their computational cost and memory consumption present significant challenges for scalability to massive, state-wide networks, reinforcing the central problem our work aims to address.

\noindent\textbf{Position Embedding.}
Position embeddings are crucial in natural language processing as they enable models to recognize the order of tokens within a sequence. In transformer architectures~\cite{vaswani2017attention}, self-attention is position agnostic, which requires position embedding to distinguish different tokens. 
It leverages sinusoidal position encodings to distinguish between tokens in a sequence.
Other works learn a separate position embedding for different positions~\cite{kenton2019bert,gehring2017convolutional}, which is more flexible and adaptive based on different tasks compared to manual embeddings like sinusoidal ones.
However, these embeddings need to pre-determine max sequence length, and hence are not suitable for graph-based tasks, where node numbers are variable~\cite{chu2021conditional,liu2020learning}.
Latter works find that such position-aware information will be lost through layers and add position embeddings at every layer of transformers~\cite{al2019character,dehghani2018universal,guo2019low,liu2020learning}.
Another line of work~\cite{shaw2018self,huang2018music,dai2019transformer,wang2021peonbert,shaw2018relative,su2023roformer} is relative position embedding, focusing on relative positions between tokens, instead of the absolute position of tokens in a sequence.
However, in graph scenarios, the relative positional information is encoded by graph operators and the missed embedding feature is the absolute one, which prompts us to leverage absolute position embedding in our model.

\section{3 Preliminaries}
\textbf{Traffic Graph Signal Representation.} The traffic network is modeled as a graph $G = (V, E, A)$, where $V$ consists of $N$ traffic flow sensors (nodes), $E$ represents edges between two sensors, and graph adjacency matrix $\mathbf{A} \in \mathbb{R}^{N\times N}$ is constructed by pairwise geodesic distances among sensors. At any time $t$, the graph signal is captured by $\mathbf{X}_t \in \mathbb{R}^{N \times D}$, where $D$ includes traffic flow, alongside temporal features (day of the week, time of day) and spatial characteristics (number of lanes, highway categories). The series of graph signals over $T$ time slices is encoded by a three-way tensor $\mathbf{X} = (\mathbf{X}_1, \mathbf{X}_2, \ldots, \mathbf{X}_T) \in \mathbb{R}^{T \times N \times D}$, visualized in Figure~\ref{fig:framework}(a).

\noindent\textbf{Traffic Flow Prediction.} The objective of this study is to predict future traffic flow by leveraging historical graph signal data. Utilizing the tensor $\mathbf{X}$ derived from network $G$, we aim to learn a function $f$ that projects the traffic flow from the past $T$ slices into the subsequent $T'$ slices, 
\begin{equation} \label{eq:prediction}
[\mathbf{X}_{(t-T+1)}, \ldots, \mathbf{X}_t; G] \xrightarrow{f} [\mathbf{X}_{(t+1)}, \ldots, \mathbf{X}_{(t+T')}].
\end{equation}
Our work focuses on architectural innovations in spatiotemporal computational models, designed to enhance efficiency and scalability in large-scale traffic prediction using standardized data formats. Factors beyond the historical traffic flow in time series and the sensor relationships represented in the graph structure are not within the scope of this study.

\section{4 Methodology}
\begin{figure}[htbp]
  \centering
  \includegraphics[width=1\textwidth]{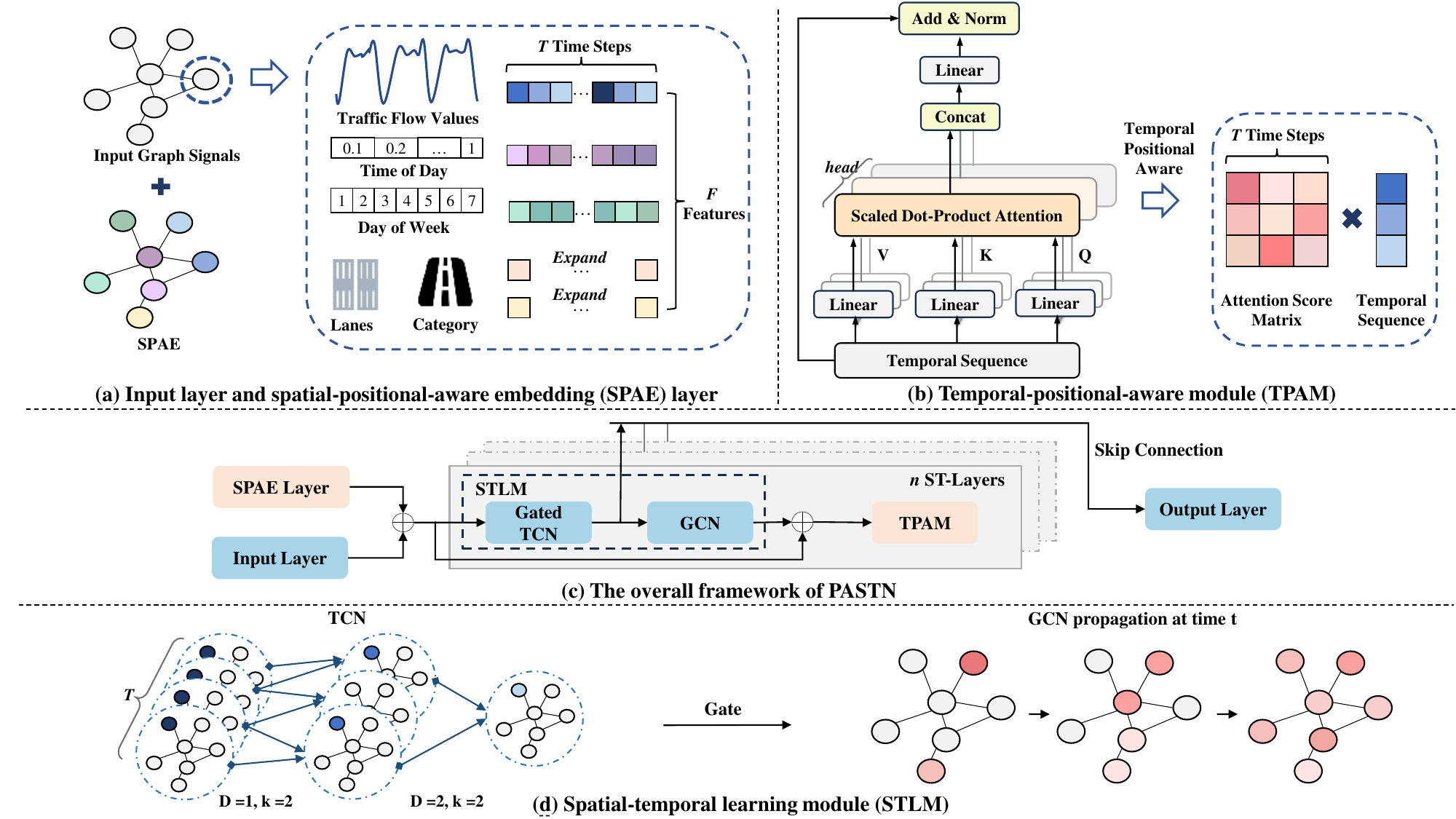}
  \caption{Detailed framework of PASTN.}
  \label{fig:framework}
\end{figure}

To address the challenges of dynamic spatial and temporal variability in traffic flow data, our proposed model, PASTN, is designed to enhance node differentiation and optimize time-step communication. The model begins with an Input Layer that uses a linear convolution layer for data preprocessing. Subsequently, the Spatial-Positional-Aware Embedding Layer (SPAE-Layer) is engineered to capture unique identity encodings for each sensor~(\S\ref{spatial_embedding}). \S\ref{backbone} describes the stacked Spatio-temporal  Layers (ST-Layers), each composed of a fundamental Spatio-temporal  Learning Module (STLM) followed by a Temporal-Positional-Aware Module (TPAM). TPAM is designed to strengthen the informational exchange across different time steps based on an attention-based mechanism~(\S\ref{temporal_attention}). Finally, the output of the ST-Layers, integrated via skip connections, feeds into the Output Layer, which contains two linear convolution layers. Skip connections address the challenges of vanishing gradients in deep architectures by enabling layers to approximate identity mappings when optimal~\cite{he2016deep,huang2017densely}, allowing the ST-Layers to capture  spatiotemporal features effectively across multiple layers.
The framework of PASTN is presented in Figure~\ref{fig:framework}. 

\subsection{4.1 Spatial-Positional-Aware Embedding}
\label{spatial_embedding}
Graph over-smoothing in multi-layer GNNs leads to feature vectors of nodes becoming indistinguishable after multiple propagation layers. This effect intensifies in large-scale networks with an exponential increase in the node number and denser connectivity, resulting in feature homogenization that impedes distinct node identification~\cite{rusch2023survey}, which will be presented in \S\ref{example}. To address this, we propose a spatial-positional-aware embedding (SPAE) method that leverages learnable spatial positional encodings to enhance node distinguishability. These embeddings are dynamically updated during training, assigning a unique encoding to each node to improve the differentiation of nodes over multiple layers of propagation. Inspired by the encoding strategies in the Caltrans Performance Measurement System, SPAE utilizes positional encodings initialized by a sinusoidal and cosine function-based method. This initialization captures both the physical and logical distances among sensors efficiently, preserving their relative positions. The initial encoding is defined as:
\begin{align}
SPAE(i, k) = \begin{cases} 
\sin\left(\frac{i}{10000^{2k/d_{model}}}\right) & \text{if } k \mod 2 = 0 \\
\cos\left(\frac{i}{10000^{2k/d_{model}}}\right) & \text{if } k \mod 2 \neq 0,
\end{cases}
\end{align}
where \( i \) represents the position of the sensor, \( k \) denotes the dimension index, and \( d_{model} \) is the dimensionality of the input. The positional encodings are directly added to the input layer, following a widely recognized approach in sequence modeling architectures, including Transformer~\cite{vaswani2017attention} and BERT~\cite{kenton2019bert}. This design effectively embeds learnable spatial relationships into the model while preserving the input dimensionality, which is consistent with the lightweight and computationally efficient principles of our proposed framework.

\subsection{4.2 Spatiotemporal  Learning Module}
\label{backbone}
Similar to existing methods such as GraphWavenet~\cite{wu2019graph}, we adopt graph convolution~\cite{defferrard2016convolutional,kipf2017semi} and linear convolution~\cite{yu2015multi} to capture the spatial and temporal correlations among sensors respectively. Specifically, the spatial dependencies are learned using both predefined and self-learned graph structures, while temporal dependencies are addressed using a TCN with a dilation factor to extend the receptive field efficiently. The graph convolution can be formulated as:
\begin{equation}
\mathbf{Z} = \sum_{k=0}^{K} \mathbf{P}_f^k \mathbf{X_{t}} \mathbf{W}_{k1} + \mathbf{P}_b^k \mathbf{X_{t}} \mathbf{W}_{k2} + \widetilde{\mathbf{A}}_{apt}^k \mathbf{X_{t}} \mathbf{W}_{k3},
\end{equation}

\noindent where $\mathbf{X_{t}} \in \mathbb{R}^{N \times D}$ represents the input graph signals at time $t$, and $\mathbf{Z} \in \mathbb{R}^{N \times M}$ denotes the output. Here, $\mathbf{W}_{k1}$, $\mathbf{W}_{k2}$, and $\mathbf{W}_{k3} \in \mathbb{R}^{D \times M}$ are the parameter matrices corresponding to the different graph convolutions, and $\widetilde{\mathbf{A}}_{apt}^k \in \mathbb{R}^{N \times N}$ is the self-learned adjacency matrix. The forward transition matrix $\mathbf{P}_f$ and backward transition matrix $\mathbf{P}_b$ are derived from the real-world normalized adjacency matrix $\mathbf{A}$, calculated as $\mathbf{P}_f = \frac{\mathbf{A}}{\text{rowsum}(\mathbf{A})}$ and $\mathbf{P}_b = \frac{\mathbf{A}^T}{\text{rowsum}(\mathbf{A}^T)}$, respectively.

For capturing longer-range temporal dependencies with lower computational burden, a TCN with a dilation factor $D$ is utilized~\cite{yu2015multi}. Mathematically, given an input sequence \(\mathbf{x} \in \mathbb{R}^T\) and a filter \(\mathbf{f} \in \mathbb{R}^k\), the convolution operation is defined as:
\begin{equation}
\mathbf{x} * \mathbf{f}(t) = \sum_{s=0}^{k-1} \mathbf{f}(s) \mathbf{x}(t - D \times s).
\end{equation}

To further enhance the model’s capability in handling complex temporal patterns, a gated TCN structure~\cite{dauphin2017language} is adopted to adaptively manage the contributions of different temporal segments. Here, $\mathbf{h}$ represents the output of the gated TCN before transmission to the graph convolution, specifically:

\begin{equation}
\mathbf{h} = tanh(\Theta_1 * \mathbf{X} + \mathbf{b}) \odot \sigma(\Theta_2 * \mathbf{X} + \mathbf{c}),
\end{equation}
\noindent with $\mathbf{X} \in \mathbb{R}^{N \times D \times T}$ as the input. In this expression, $\Theta_1$ and $\Theta_2$ are the model parameters for the gating mechanisms, $\mathbf{b}$ and $\mathbf{c} \in \mathbb{R}^{D}$ are bias terms, $tanh(\cdot)$ represents the tangent hyperbolic function, and $\sigma(\cdot)$ denotes the sigmoid function. Figure~\ref{fig:framework}(d)
shows how graph convolution and TCN collaborate with each other.

\subsection{4.3 Temporal-Positional-Aware Module}
\label{temporal_attention}
The relationship between traffic conditions at different times is dynamic, exhibiting trends and periodicities that vary across different situations. To strengthen the capture of evolving temporal patterns, we introduce the Temporal-Positional-Aware Module (TPAM), as illustrated in Figure~\ref{fig:framework}(b). Formally, for a given node $n$, we first obtain the query, key, and value matrices as:
\begin{equation}
\mathbf{Q}^{T}_n = \mathbf{X}_{n} \mathbf{W}_Q^T, \quad \mathbf{K}^{T}_n = \mathbf{X}_{n} \mathbf{W}_K^T, \quad \mathbf{V}^{T}_n = \mathbf{X}_{n} \mathbf{W}_V^T,
\end{equation}
where $\mathbf{W}_Q^T, \mathbf{W}_K^T, \mathbf{W}_V^T \in \mathbb{R}^{D \times d}$ are learnable parameters.

For multi-head attention, each matrix is directly split into segments corresponding to each head:
\begin{equation}
\mathbf{Q}^{T, h}_n = \text{split}(\mathbf{Q}^{T}_n, h), \quad \mathbf{K}^{T, h}_n = \text{split}(\mathbf{K}^{T}_n, h),\quad \mathbf{V}^{T, h}_n = \text{split}(\mathbf{V}^{T}_n, h),
\end{equation}
where $\text{split}(\cdot, h)$ denotes the operation of dividing the matrix into $h$ smaller matrices, each representing a head’s portion of the original matrix dimensions. The multi-head architecture here allows the model to capture diverse predictive patterns by having each head specialize in attending to information from a different representation subspace, yielding a more internally consistent and structured multi-step prediction.

Then, we apply multi-head self-attention~\cite{vaswani2017attention} operations in the temporal dimension and obtain the temporal dependencies between all time slices for node $n$ as:
\begin{equation}
\mathbf{A}^{T, h}_n = \frac{\mathbf{Q}^{T, h}_n (\mathbf{K}^{T, h}_n)^T}{\sqrt{d_h}},
\end{equation}
where $\mathbf{A}^{T, h}_n$ is the attention score matrix for each head and \( d_h \) is the dimensionality of each head.

It can be seen that temporal self-attention has a global receptive to model the dynamic long-range temporal dependencies among all time slices. The attention scores are further processed through a softmax function to ensure normalization, after which the outputs of each head are concatenated and linearly transformed:
\begin{equation}
\mathrm{MultiheadAttention}(\mathbf{Q}_n, \mathbf{K}_n, \mathbf{V}_n) = \bigg[\bigoplus_{h=1}^{H} \mathrm{softmax}(\mathbf{A}_{n,h}) \mathbf{V}_{n,h}\bigg] \mathbf{W}_O,
\end{equation}
where $\mathbf{W}_O \in \mathbb{R}^{d \times D}$ is a learnable matrix, and $\bigoplus$ denotes concatenation.

Finally, we integrate the original features with the attention-enhanced features through a residual connection~\cite{he2016deep}, followed by layer normalization~\cite{xu2019understanding} to obtain the output of TPAM:
\begin{equation}\label{eq:TPAM}
\text{TPAM}= \mathrm{LayerNorm}(\mathrm{MultiheadAttention}(\mathbf{Q}^{T}_n, \mathbf{K}^{T}_n, \mathbf{V}^{T}_n)+\mathbf{X}_{n})
\end{equation}

\subsection{4.4 Loss Function}
Mean absolute error (MAE) is chosen as the training objective defined by:
\begin{equation}
L(\hat{\mathbf{X}}_{(t+1):(t+T')}, \Theta) = \frac{1}{T'N} \sum_{i=1}^{T'} \sum_{j=1}^{N} \left| \hat{\mathbf{X}}_{j}^{(t+i)} - \mathbf{X}_{j}^{(t+i)} \right|,
\end{equation}
where $\hat{\mathbf{X}}_{(t+1):(t+T')}$ is the predicted traffic flow values across $T'$ time steps, $\Theta$ denotes all learnable model parameters, and $\mathbf{X}_{j}^{(t+i)}$ denotes the ground truth.

\section{5 Experiments}
\subsection{5.1 Experiment Settings}
In this section, we evaluate our PASTN model using four real-world traffic flow datasets focusing on large scales, including datasets from counties, megalopolis areas, and state-level networks, as detailed in Table~\ref{tab:datasets}. The datasets are split chronological into train, validation, and test sets, with a ratio of 6:2:2.

\begin{table}[htbp]
\setlength{\tabcolsep}{3pt}

\centering
        \begin{tabular}{l l l l l}
            \toprule
            Dataset & Traffic Scale & Nodes & Edges & Time Range \\
            \hline
            SD: San Diego County  & County & 716 & 17,319 & 1/Jan/2017 - 31/Dec/2021 \\
            GBA: Greater Bay Area & Megalopolis & 2,352 & 61,246 & 1/Jan/2017 - 31/Dec/2021\\
            GLA: Greater Los Angeles Area  & Megalopolis & 3,834 & 98,703 & 1/Jan/2017 - 31/Dec/2021 \\
            CA: California State & State & 8,600 & 201,363 & 1/Jan/2017 - 31/Dec/2021 \\
            \bottomrule
        \end{tabular}
        \caption{Datasets originate from the LargeST dataset collection~\cite{liu2023largest}.
            The sampling rate for all datasets is at 5-minute-level. Greater Bay Area (GBA) includes Alameda, Contra Costa, Marin, Napa, San Benito, San Francisco, San Mateo, Santa Clara, Santa Cruz, Solano, and Sonoma. Greater Los Angeles Area (GLA) encompasses Los Angeles, Orange, Riverside, San Bernardino, and Ventura.}\label{tab:datasets}

\end{table}

\noindent\textbf{Baseline.} We evaluate the performance of our model against 12 leading traffic forecasting baselines, with the standard setting of predicting the 12-step future values based on the 12-step historical data~\cite{li2018diffusion,wu2019graph}. 
We compare our model with temporal-only method, LSTM~\cite{hochreiter1997long}, temporal-spatial model leveraging GNN and RNN, including DCRNN~\cite{li2018diffusion}, AGCRN~\cite{bai2020adaptive}, TCN-based methods like STGCN~\cite{yu2018spatio} and GWNET~\cite{wu2019graph}, and attention-based methods including ASTGCN~\cite{guo2019attention}, STTN~\cite{xu2020spatial}, and PDformer \cite{jiang2023pdformer}.
Furthermore, we include four cutting-edge methods that reflect current research trends in the field: STGODE~\cite{fang2021spatial}, DSTAGNN~\cite{lan2022dstagnn}, DGCRN~\cite{li2021dynamic}, and $\text{D}^{2}\text{STGNN}$~\cite{shao2022decoupled}.

\noindent\textbf{Model Settings.} Experiments are conducted on a computing server equipped with an Intel(R) Xeon(R) Gold 6248R CPU @ 3.00 GHz, 503 GB RAM and an NVIDIA GeForce 3090 GPU with 24 GB memory. 
Each experiment is independently repeated twice for 100 epochs, and the average performance is reported. 
For model and training configurations, we adhere to the recommended settings suggested by~\citet{liu2023largest}. 
We investigate the architecture of ST-layers with depths of \{4, 6, 8\}. TCN modules in the STLM have fixed residual channels, explored over \{16, 24, 32\}, while the dilation factors $D$ alternate between 1 and 2 across the sequence of model layers. The number of graph propagation layers is examined over \{1, 2, 3\}. 
Additionally, for the TPAM, the number of attention heads is searched over \{2, 4, 8\}. These values were carefully chosen to stay within the GPU memory constraints, as larger configurations result in out-of-memory issues. Our experiments also showed that the best performance was achieved at the upper bounds of these configurations.  We train our model by Adam optimizer~\cite{KingBa15} with an initial learning rate of 0.001. Dropout with p = 0.3 is applied to the outputs of the graph convolution modules. 

\noindent\textbf{Evaluation Metrics.} We perform a comprehensive comparison of PASTN with the baselines from the following perspectives: (1) \textit{Performance:} Three common metrics are adopted: Mean Absolute Error (MAE), Root Mean Square Error (RMSE), and Mean Absolute Percentage Error (MAPE). (2) \textit{Efficiency:} We report the number of learnable parameters and the training time to reflect the model's complexity. For inference efficiency, we report the total wall-clock time required to process the entire validation set, which consists of 21,024 time points. The values reported in Table~\ref{tab:efficiency} represent the cumulative time, where each time point involves a single inference pass over the entire network. All models achieve inference times well below the data update interval of 5 minutes, with each inference for the entire network completing in less than 0.05 seconds, as shown in Table~\ref{tab:efficiency}. Models with shorter inference times, however, demonstrate greater potential for deployment in time-sensitive applications, such as adaptive traffic control and navigation systems.

\subsection{5.2 Effectiveness}
\textbf{Overall Performance.} Our analysis draws on traffic data from 2019 across multi-scale datasets, as detailed in Table~\ref{tab:overall performance}. Our model, PASTN, outperforms all other approaches, particularly in challenging large-scale predictions. 
Noticeably, recent models like ASTGCN and DSTAGNN face challenges across all datasets, possibly due to their high computational complexity, which can lead to severe overfit problems. 
Meanwhile, DGCRN and $\text{D}^{2}\text{STGNN}$, although effective at capturing dynamic spatial topology in smaller datasets, struggle to scale to larger datasets. 
In contrast, our PASTN model excels on all genres of datasets, especially on the state-level CA dataset, with substantial improvements of 10.98\% in MAE, 11.23\% in MAPE, and 18.45\% in RMSE, proving its efficiency in handling large-scale traffic data.

\begin{table}[htbp]
\resizebox{\textwidth}{!}{%

    \begin{tabular}{cccccccccccccc}
    \toprule
    \multirow{3}[6]{*}{\textbf{Type}} & \multicolumn{1}{l}{\multirow{3}[6]{*}{\textbf{Model}}} & \multicolumn{3}{c}{\textbf{County}} & \multicolumn{6}{c}{\textbf{Megalopolis}}     & \multicolumn{3}{c}{\textbf{State}} \\
\cmidrule{3-14}          &       & \multicolumn{3}{c}{\textbf{SD}} & \multicolumn{3}{c}{\textbf{GBA}} & \multicolumn{3}{c}{\textbf{GLA}} & \multicolumn{3}{c}{\textbf{CA}} \\
\cmidrule{3-14}          &       & \textbf{MAE} & \textbf{RMSE} & \textbf{MAPE} & \textbf{MAE} & \textbf{RMSE} & \textbf{MAPE} & \textbf{MAE} & \textbf{RMSE} & \textbf{MAPE} & \textbf{MAE} & \textbf{RMSE} & \textbf{MAPE} \\
    \midrule
    Temporal-only & \multicolumn{1}{l}{LSTM*} & 26.43 & 41.72 & 17.19\% & 27.97 & 44.21 & 24.48\% & 28.06 & 44.38 & 17.24\% & 26.89 & 43.11 & 20.16\% \\
    \midrule
    \multirow{2}[2]{*}{RNN-based} & \multicolumn{1}{l}{DCRNN*} & 21.04 & 33.35 & 14.13\% & 23.13 & 36.35 & 20.84\% & 23.17 & 36.19 & 14.40\% & 21.89 & 34.41 & 17.06\% \\
          & \multicolumn{1}{l}{AGCRN*} & 18.09 & 32.04 & 13.28\% & 21.03 & 34.26 & 16.90\% & 20.23 & 34.84 & 12.87\% & -     & -     & - \\
    \midrule
    \multirow{2}[2]{*}{TCN-based} & \multicolumn{1}{l}{STGCN*} & 19.69 & 34.14 & 13.86\% & 23.42 & 38.57 & 18.47\% & 22.64 & 38.81 & 14.18\% & 21.35 & 36.39 & 16.53\% \\
          & \multicolumn{1}{l}{GWNET*} & 17.74 & 29.64 & 11.88\% & 20.91 & 33.41 & 17.66\% & 21.20 & 33.59 & 13.18\% & 21.72 & 34.22 & 17.40\% \\
    \midrule
    \multirow{3}[2]{*}{Attention-based} & \multicolumn{1}{l}{ASTGCN} & 23.70 & 37.63 & 15.65\% & 26.47 & 40.99 & 23.65\% & 28.99 & 44.33 & 19.62\% & -     & -     & - \\
          & \multicolumn{1}{l}{STTN} & 18.69 & 31.11 & 12.82\% & 20.97 & 33.78 & 16.84\% & -     & -     & -     & -     & -     & - \\
          & \multicolumn{1}{l}{PDFormer} & 19.97 & 33.51 & 13.97\% & -     & -     & -     & -     & -     & -     & -     & -     & - \\
    \midrule
    ODE-based & \multicolumn{1}{l}{STGODE*} & 19.55 & 33.57 & 13.24\% & 21.79 & 35.37 & 18.26\% & 21.49 & 36.14 & 13.73\% & 20.78 & 36.60 & 16.81\% \\
    \midrule
    \multirow{3}[2]{*}{Dynamic graph} & \multicolumn{1}{l}{DSTAGNN} & 21.82 & 34.68 & 14.40\% & 23.82 & 37.29 & 20.16\% & 24.13 & 38.15 & 15.07\% & -     & -     & - \\
          & \multicolumn{1}{l}{DGCRN*} & 18.03 & 30.09 & 12.09\% & 20.91 & 33.84 & 16.88\% & -     & -     & -     & -     & -     & - \\
          & \multicolumn{1}{l}{$\text{D}^{2}\text{STGNN}$} & 17.85 & 29.51 & 11.54\% & 20.71 & 33.65 & 15.04\% & -     & -     & -     & -     & -     & - \\
    \midrule
    \rowcolor[rgb]{ .929,  .929,  .929} \multicolumn{2}{c}{PASTN} & \textbf{17.32} & \textbf{28.94} & \textbf{11.46\%} & \textbf{20.13} & \textbf{32.98} & \textbf{14.78\%} & \textbf{19.86} & \textbf{32.29} & \textbf{12.52\%} & \textbf{18.49} & \textbf{30.36} & \textbf{13.48\%} \\

    \bottomrule
    \end{tabular}%
    }
    \caption{Performance comparison across different traffic scales. Results for models marked with * were obtained from our reproductions. Other baseline results are cited from previous works~\cite{liu2023largest, wang2024spatiotemporal} utilizing an A6000 GPU with 48 GB memory. ``-'' indicates that the models incur out-of-memory issues. Values in bold indicate the best.}
    
  \label{tab:overall performance}%
\end{table}

\begin{figure}[ht]
  \centering
  \includegraphics[width=0.7\textwidth]{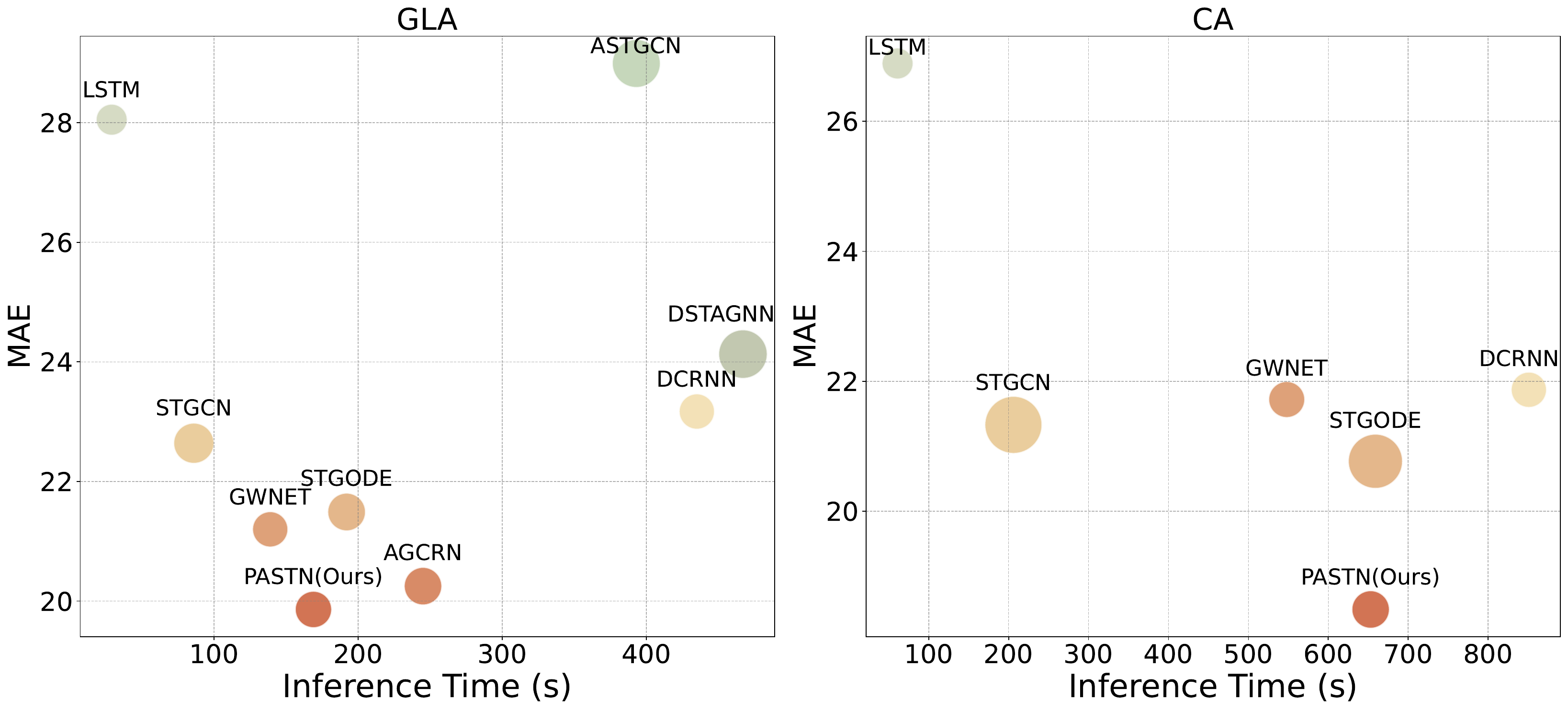}
  \caption{Performance \textit{vs.} inference time on GLA and CA datasets. The bubble size represents the learnable parameter volume.}
\label{fig:bubble}
\end{figure}

\noindent\textbf{Model Efficiency.} 
To comprehensively evaluate our method in terms of both efficacy and efficiency, we compare each model from following perspectives: the number of tunable parameters, training time, inference latency as well as performance.
We summarize the efficiency comparisons in Table~\ref{tab:efficiency} with the following observations. 
TCN-based approaches like STGCN and GWNET are generally faster attributed to the parallel processing capabilities of temporal convolution operations.  
In contrast, models like DGCRN and $\text{D}^{2}\text{STGNN}$, despite strong performance, are hindered by longer inference times and non-scalability to large networks.
Moreover, significant variations are observed in the number of learnable parameters across models. 
For instance, in evaluating performance on the CA dataset, we find that while STGCN and STGODE exceed GWNET in MAE/MAPE, they require substantially more parameters, ranging from double to ten times those of GWNET. 
For a fair assessment, 
Figure~\ref{fig:bubble} provides a comprehensive visualization of both performance and efficiency on the larger dataset: GLA and CA, demonstrating that PASTN achieves highly competitive performance-efficiency trade-off among SOTA methods.
\begin{table}[htbp]
\setlength{\tabcolsep}{3pt} 
\renewcommand{\arraystretch}{0.8} 
\centering

\begin{adjustbox}{center}
\scriptsize

\begin{tabular}{ccccccccccccc}
\toprule
\multirow{3}{*}{\textbf{Model}}  &
 \multicolumn{3}{c}{\textbf{County}} & \multicolumn{6}{c}{\textbf{Megalopolis}} & \multicolumn{3}{c}{\textbf{State}} \\
 \cmidrule(lr){2-13} 
  &  \multicolumn{3}{c}{\textbf{SD}} & \multicolumn{3}{c}{\textbf{GBA}} & \multicolumn{3}{c}{\textbf{GLA}} & \multicolumn{3}{c}{\textbf{CA}}\\
  \cmidrule(lr){2-13} 
 & Parameters & Train & Infer & Parameters & Train & Infer &Parameters & Train & Infer &Parameters & Train & Infer  \\
\midrule
\multicolumn{1}{c}{LSTM}	&98K	&28 &6	&98K	&128 &17	&98K	&209	&29 &98K	&487 &61\\
\multicolumn{1}{c}{DCRNN}	&373K	&1068 &150	&373K	&2616 &319	&373K	&3284 &435	&373K	&5812 &851\\
\multicolumn{1}{c}{AGCRN}	&761K	&122 &15	&777K	&638 &83	&792K	&1913 &245	&-	&- &-\\
\multicolumn{1}{c}{STGCN}	&508K	&73&16	&1.3M	&219&54	&2.1M	&386&86	&4.5M	&902&206\\
\multicolumn{1}{c}{GWNET}	&311K	&104&14	&344K	&513&66	&374K	&1128&139	&469K	&4250&548\\
\multicolumn{1}{c}{ASTGCN}	&2.2M	&-&19	&22.3M	&-&147	&51.9M	&-&393	&-	&-&-\\
\multicolumn{1}{c}{STTN}	&114K	&-&26	&218K	&-&197	&-	&-	&-	&-&-\\
\multicolumn{1}{c}{PDFormer}	&3.7M	&-&324	&-	&- &- &-	&-	&-	&-	&-\\
\multicolumn{1}{c}{STGODE}	&729K	&208 &26	&788K	&759 &103	&841K	&1505 &192	&1.0M	&4412&659\\
\multicolumn{1}{c}{DSTAGNN}	&3.9M	&-&23	&26.9M	&-&171	&66.3M	&-&467	&-	&-&-\\
\multicolumn{1}{c}{DGCRN}	&243K	&480&76	&374K	&5461&605	&-	&-	&-	&- &-	&-\\
\multicolumn{1}{c}{$\text{D}^{2}\text{STGNN}$}	&406K	&- &69	&446K	&- &796	&-	&-	&-	&- &-	&-\\
\rowcolor[rgb]{ .929,  .929,  .929} \multicolumn{1}{c}{PASTN}	&368K	&118 &18	&453K	&528 &81	&530K	&1210 &169	&778K	&4313 &653\\
\bottomrule

\end{tabular}

\end{adjustbox}
\caption{Efficiency comparison. Parameters:  the number of learnable parameters (K:$10^3$, M:$10^6$). Train: training time (s) per epoch. Infer: inference time (s) on the validation set. All metrics are reported for models we reproduced. For other baselines cited from~\cite{liu2023largest, wang2024spatiotemporal}, we omit training time ('–') as a direct hardware comparison is misleading.}

\label{tab:efficiency}
\end{table}

\subsection{5.3 Generalization}
To explore the generalization capability of PASTN, we conduct experiments on datasets of different years: 2018 and 2020. The year 2018 represents a pre-COVID-19 scenario, while 2020 data reflects the impacts of the pandemic. This selection is crucial to understand the scalability of our models under varying social-economic conditions. Selected baselines are within our computational budgets. The results, as shown in Table~\ref{tab:generalization}, indicate that the performance trends among the models are consistent with previous findings in 2019 data, with PASTN outperforming the baselines in both regions and years studied. Interestingly, all models achieved notably better performance in 2020. The improvement can be partially attributed to the governmental responses to the pandemic, including stay-at-home orders, social distancing mandates, school closures, and travel restrictions, which significantly altered mobility patterns~\cite{mcelroy2022understanding,hall2023inferred,bhagat2023rush}. These actions likely led to a reduction in traffic variability and complexities, possibly making traffic flows more predictable and consistent across different times and zones. To provide deeper insights into the adaptability of traffic prediction models under such disruptive global events, a micro-level analysis is required to explore the correlations between the pandemic's impact on travel behaviors in further studies.
\begin{table}[htbp]
\renewcommand{\arraystretch}{0.8}
  \centering
\resizebox{\textwidth}{!}{%
\scriptsize
    \begin{tabular}{ccccccccccccc}
    \toprule
    \multicolumn{1}{c}{\multirow{2}[4]{*}{\textbf{Model}}} & \multicolumn{3}{c}{\textbf{GBA 2018}} & \multicolumn{3}{c}{\textbf{GLA 2018}} & \multicolumn{3}{c}{\textbf{GBA 2020}} & \multicolumn{3}{c}{\textbf{GLA 2020}} \\
\cmidrule{2-13}          & \textbf{MAE} & \textbf{RMSE} & \textbf{MAPE} & \textbf{MAE} & \textbf{RMSE} & \textbf{MAPE} & \textbf{MAE} & \textbf{RMSE} & \textbf{MAPE} & \textbf{MAE} & \textbf{RMSE} & \textbf{MAPE} \\
    \midrule
    DCRNN & 25.04 & 37.45 & 23.36\% & 24.23 & 37.87 & 14.56\% & 20.12 & 31.07 & 16.23\% & 21.85 & 32.59 & 16.34\% \\
    AGCRN & 22.7  & 38.76 & 21.60\% & 23.42 & 39.12 & 15.14\% & 15.36 & 25.82 & 11.16\% & 18.43 & 31.92 & 15.61\% \\
    STGCN & 25.73 & 41.56 & 24.68\% & 26.31 & 43.56 & 17.18\% & 17.79 & 29.72 & 13.25\% & 21.82 & 36.37 & 17.99\% \\
    GWNET & 22.94 & 36.83 & 22.77\% & 23.43 & 36.47 & 13.83\% & 16.18 & 26.55 & 12.00\% & 20.28 & 32.66 & 14.66\% \\
    STGODE & 22.94 & 38.12 & 21.78\% & 24.02 & \textbf{29.75} & 15.96\% & 17.54 & 28.77 & 12.77\% & 19.14 & 32.01 & 15.92\% \\
    \rowcolor[rgb]{ .929,  .929,  .929} PASTN  & \textbf{21.94} & \textbf{35.44} & \textbf{19.38\%} & \textbf{21.76} & 34.84 & \textbf{13.13\%} & \textbf{14.96} & \textbf{25.17} & \textbf{11.04\%} & \textbf{18.27} & \textbf{30.04} & \textbf{13.15\%} \\
    \bottomrule
    \end{tabular}%
    }
      \caption{Generalization results of our method in terms of datasets and years. }
  \label{tab:generalization}%
\end{table}%

\subsection{5.4 Ablations}
\label{ablation}
To further investigate the effectiveness of components in
PASTN, our ablation studies are organized into following variants: (1) \textit{w/o SPAE:} Tests PASTN without the SPAE layer; (2) \textit{w/o TPAM:} Examines PASTN minus the TPAM; (3) \textit{ST:} Utilizes only the spatio-temporal backbone.  (4) \textit{w/ RI:} Implements random initialization to determine if enhancements result from reducing signal noise in graph learning~\cite{chen2020measuring}. (5) \textit{w/ LF:} Utilizes the encoding method where only the frequency of each node dimension is learnable~\cite{wang2021peonbert}, investigating the effectiveness of this alternative method for absolute position embedding. (6) \textit{w/ RE:} Switches from absolute to relative position embedding~\cite{shaw2018relative}, examining how spatial contextual changes affect model accuracy. (7) \textit{w/ MI:} evaluates mid-network placement of SPAE; (8) \textit{w/ EI:} assesses end-network insertion of SPAE. Variants (1), (2), and (3) are component ablation studies. Variants (4)-(6) evaluate different initialization methods for the SPAE, while Variants (7) and (8) examine the impact of embedding positions within the model architecture.

Additionally, to specifically assess SPAE’s adaptability in spatial perception learning, we designed three complementary experiments: (9) \textit{w/ SP:} Sensor Position Perturbation, which introduces controlled noise to test the model’s resilience to slight positional changes; (10) \textit{w/ SR:} Sensor Position Reset, where the initial sensor positions are shuffled while preserving their original relative order, allowing us to evaluate the model's flexibility under altered spatial configurations; (11) \textit{w/ EF:} Embedding Fixation, where positional embeddings are fixed post-initialization to determine the importance of dynamic learning.

\begin{figure}[htbp]
  \centering
  \includegraphics[width=0.8\textwidth]{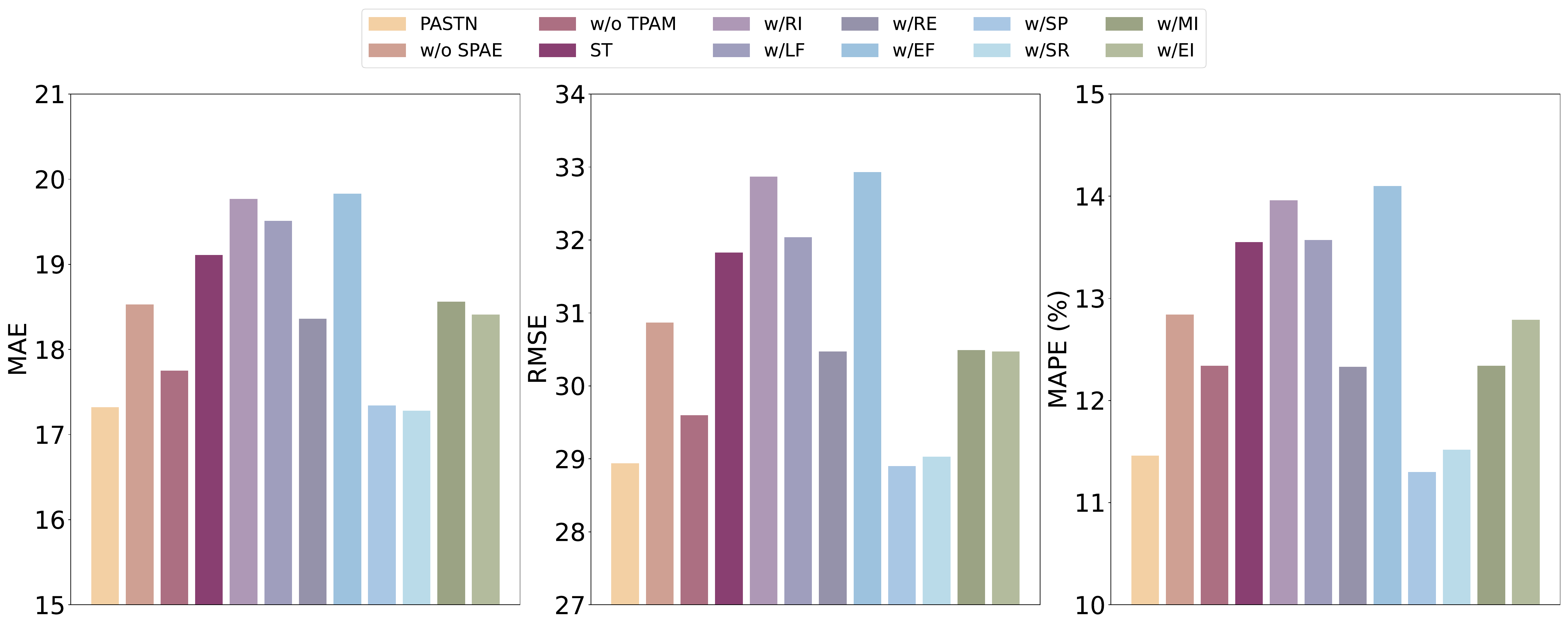}
  \caption{MAE, RMSE and MAPE on SD-19 dataset.}
\label{fig:ablation}
\end{figure}

The performance results of our ablation study are depicted in Figure~\ref{fig:ablation}. Several key observations can be made from these findings: (1) The inclusion of SPAE and TPAM significantly enhances the model's performance while the introduction of SPAE results in a more pronounced improvement, suggesting its criticality in encoding spatial information effectively. (2) The use of random initialization (\textit{w/ RI}) leads to a decline in performance, indicating that PASTN’s gains are due to structured learning mechanisms rather than noise introduction. Fixing the embeddings post-initialization (\textit{w/ EF}) further reduces performance, even below random initialization, which emphasizes the critical role of SPAE’s dynamic adaptability in spatial learning. Although relative position embedding (\textit{w/ RE}) shows some improvements, it does not surpass the efficacy of absolute position embedding. Furthermore, the \textit{w/ LF} method, which only requires learning the encoding frequency, does not perform as well as our proposed method, highlighting that a more  position-specific encoding better boosts the model's effectiveness. (3)In our additional Sensor Position Perturbation and Sensor Position Reset experiments, where sensor positions were either slightly perturbed or reset, the model’s performance remained robust. This stability indicates that SPAE is capable of capturing flexible spatial representations and adapting to varying configurations without losing effectiveness. (4)Inserting SPAE at the middle (\textit{w/ MI}) or at the end (\textit{w/ EI}) of the network yields inferior results compared to insertion at the beginning, which supports the idea that early integration of SPAE allows the network to leverage deeper and more complex spatial information.

\subsection{5.5 Case Study and Visualization}
\label{example}

\begin{figure}[htbp]
  \centering
  \includegraphics[width=0.9\textwidth]{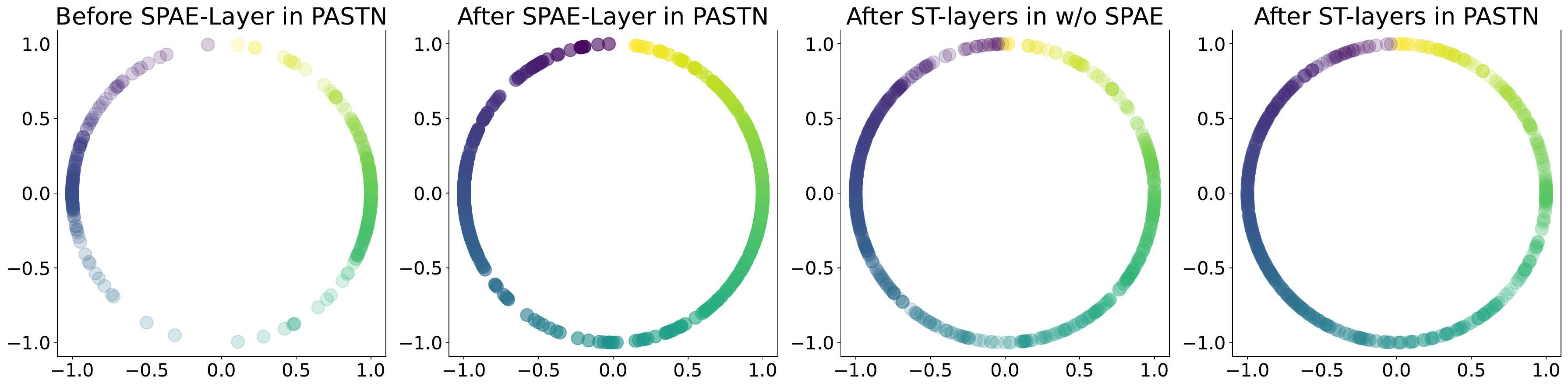}
  \caption{Embedding visualization of SPAE. A more uniform and complete circular distribution of colors indicates lower similarity among nodes.}\label{fig:SPAE}
\end{figure}

\textbf{Effect of SPAE and TPAM.} 
To assess the effectiveness of our proposed SPAE and TPAM frameworks in augmenting spatial and temporal position awareness, we visualize output embeddings from randomly selected samples within the same dataset utilized for our ablation studies (\S\ref{ablation}). For SPAE, we explore the impact of the SPAE layer by visualizing the embeddings before and after processing by it within PASTN. We further compare outputs from two variants of the PASTN model: the original model with SPAE and \textit{w/o} SPAE. Given the high-dimensional nature of the embeddings, we first apply principal component analysis (PCA) to reduce the dimensionality to 2-D~\cite{wold1987principal}. To project these 2-D embeddings onto a unit circle, we normalize each point using its L2 norm. This normalization step places each embedding point on the circumference of the unit circle, where a more uniform distribution indicates greater dispersion of node information. The resulting unit vector is denoted by:

\begin{equation}
\mathbf{Z}_{\text{unit}} = \frac{\mathbf{Z}}{\|\mathbf{Z}\|_2} = \left( \cos \theta, \sin \theta \right),
\end{equation}

\noindent where \( \mathbf{Z} = (x, y) \) is the 2-D embedding vector for each point, \( \|\mathbf{Z}\|_2 = \sqrt{x^2 + y^2} \) is its L2 norm, and \( \theta = \arctan \left( \frac{y}{x} \right) \) represents the angle of the embedding vector relative to the x-axis. As depicted in Figure~\ref{fig:SPAE}, the visualization results show that SPAE produces a more uniform distribution on the unit circle, effectively dispersing node information and mitigating the over-smoothing effect seen in \textit{w/o} SPAE.

\begin{figure}[htbp]
  \centering
  \includegraphics[width=0.7\textwidth]{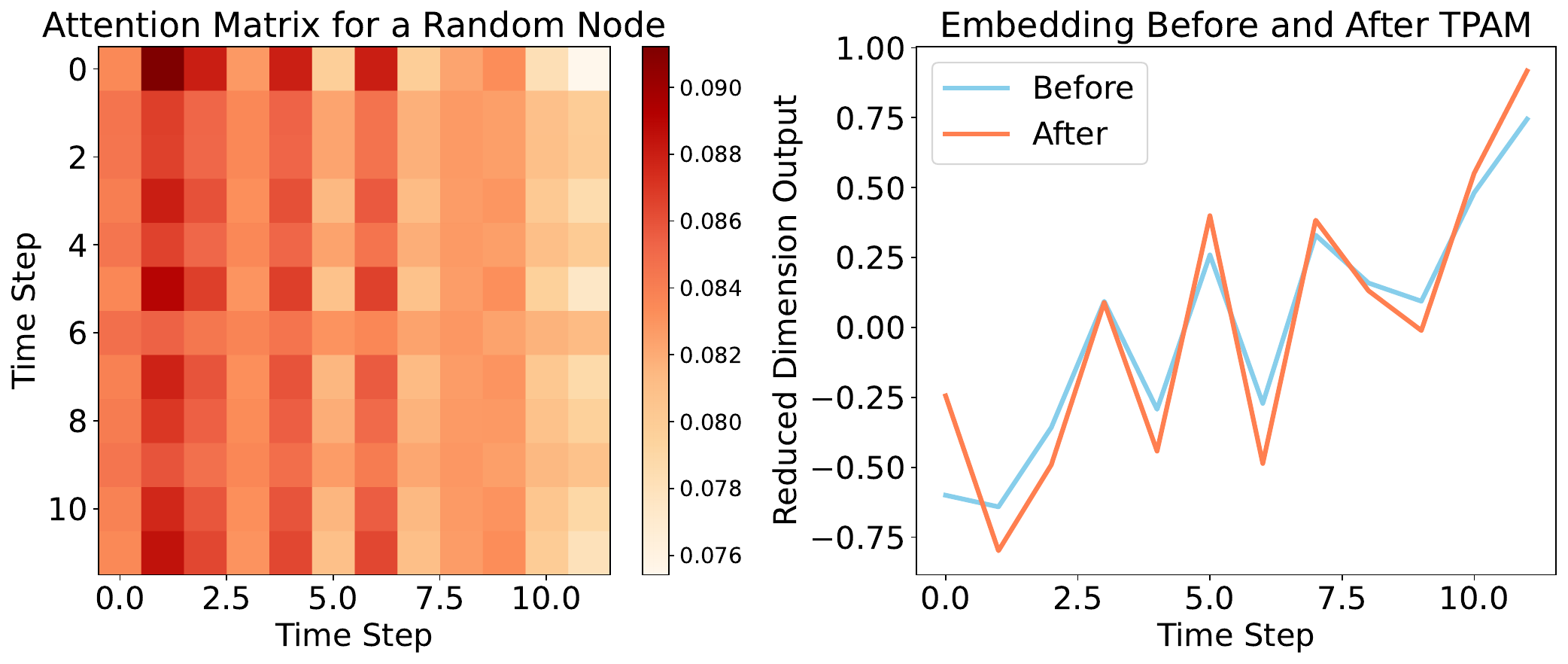}
  \caption{Attention matrix (Left) and output embedding visualization of TPAM (Right).}\label{fig:TPAM}
\end{figure}

Figure~\ref{fig:TPAM} presents the embedding of a random node before and after TPAM. Unlike traditional TCNs,  which rely on fixed kernel sizes to process only adjacent time steps in a single computation, TPAM leverages the attention mechanism to compute relationships across all time steps. The attention map generated by TPAM highlights increased focus on significant time steps within the entire sequence, reflecting the capacity to emphasize relevant temporal features. Post-TPAM, the time series exhibits more distinct trend delineation, showcasing TPAM's efficacy in capturing and understanding temporal dynamics.

\begin{figure}[htbp]
  \centering
  \includegraphics[width=0.7\textwidth]{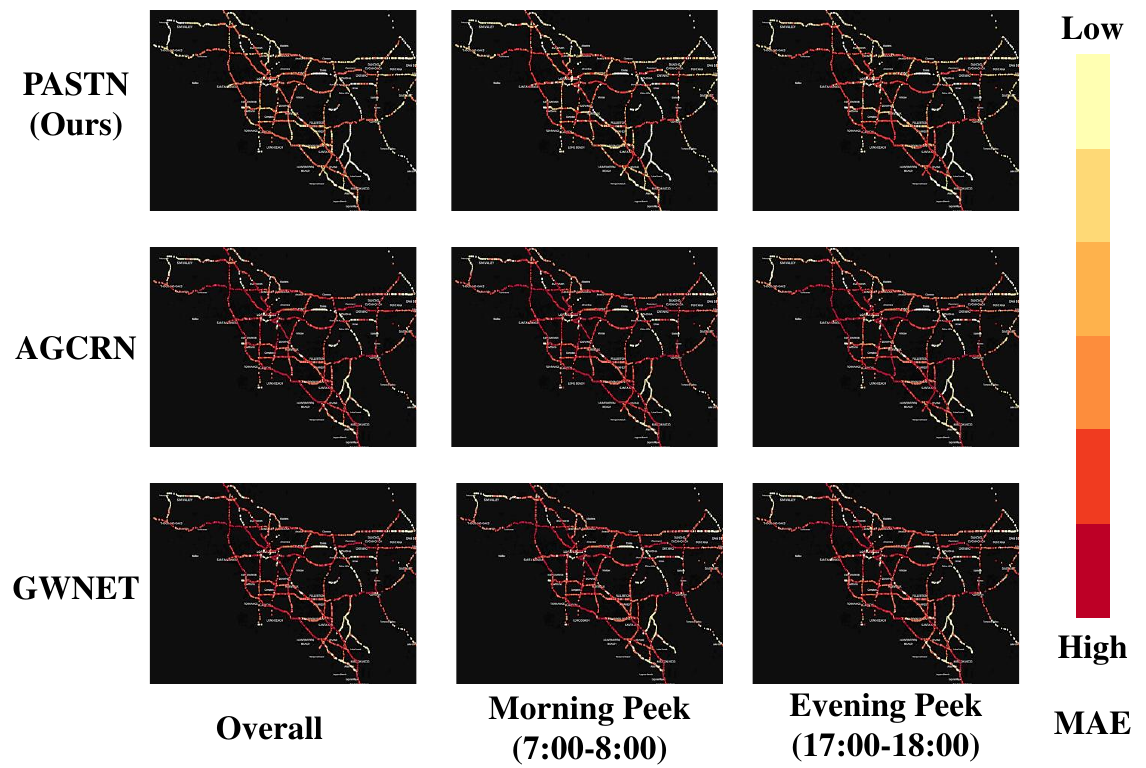}
  \caption{Cases for traffic flow forecasting in the future 1 h.}\label{fig:MAE}
\end{figure}

\noindent\textbf{Performance on Critical Regions and Times.} 
As shown in Figure~\ref{fig:MAE}, we present overall views of the prediction error for the future 1 h (12 time steps) of 3834 sensors in the GLA dataset. For this visualization, we compare PASTN against AGCRN and GWNET, which are the two strongest-performing baselines in our quantitative experiments in Table~\ref{tab:overall performance}. Overall, PASTN shows superior performance, especially in regions with complex spatio-temporal dependencies. In areas such as traffic intersections and locations near the Central Business District, where traffic uncertainty is high, the occurrence of anomalies is more frequent, resulting in lower predictability~\cite{chand2021modeling}. It is evident that predictions at such areas typically exhibit higher errors, whereas PASTN demonstrates a comparatively lower and more confined geographical range of high errors. This indicates the effectiveness of our design in enhancing GNN sensitivity to anomalies by the SPAE. Moreover, PASTN excels in capturing the time-variant dynamics of morning and evening peak periods, validating its ability to capture the evolving traffic pattern of a day. 

To evaluate its adaptability under more irregular conditions, we conducted a fine-grained analysis of holiday traffic, distinguishing between the impacts of short-term (a single day) versus long-term (an entire holiday period) distribution shifts. As summarized in Table~\ref{tab:holiday}, prediction errors during the entire holiday break are generally lower than on the specific holiday date. The results suggest that the abrupt, highly volatile traffic on the peak day itself is inherently more challenging to predict than the sustained patterns of the surrounding holiday period. PASTN consistently establishes itself as the top-performing model across all these scenarios.  Its advantage is particularly crucial on the "Single Day" peaks, which represent the more difficult prediction task for all baselines, showcasing its superior robustness against holiday traffic anomalies.

\begin{table}[htbp]
\setlength{\tabcolsep}{3.5pt} 
\renewcommand{\arraystretch}{0.9} 
\centering
\begin{adjustbox}{center}
\scriptsize
\begin{tabular}{clcccccccccccc}
\toprule
\multirow{3}{*}{Model} & \multirow{2}{*}{Holiday} & \multicolumn{6}{c}{\textbf{Christmas}} & \multicolumn{6}{c}{\textbf{Thanksgiving}} \\
\cmidrule(lr){3-8} \cmidrule(lr){9-14}
& & \multicolumn{3}{c}{\textbf{Single Day}} & \multicolumn{3}{c}{\textbf{Entire Period}} & \multicolumn{3}{c}{\textbf{Single Day}} & \multicolumn{3}{c}{\textbf{Entire Period}} \\
\cmidrule(lr){3-5} \cmidrule(lr){6-8} \cmidrule(lr){9-11} \cmidrule(lr){12-14}
& Metrics & MAE & RMSE & MAPE & MAE & RMSE & MAPE & MAE & RMSE & MAPE & MAE & RMSE & MAPE \\
\midrule
\multicolumn{2}{l}{DCRNN} & 24.29 & 38.59 & 15.11\% & 23.76 & 36.94 &  14.61\% & 23.89 & 37.86 & 14.85\% & 23.07 & 36.77 & 14.59\% \\
\multicolumn{2}{l}{AGCRN} & 21.44 & 36.92 & 13.53\% & 20.99 & 35.89 & 13.42\% & 21.05 & 36.69 & 13.31\% & 20.64& 35,76 & 12.75\% \\
\multicolumn{2}{l}{STGCN} & 23.97 & 40.21 & 15.06\% & 23.07 & 39.43 & 14.23\% & 23.37 & 39.84 & 14.53\% & 22.86 & 39.01 & 12.20\% \\
\multicolumn{2}{l}{GWNET} & 22.36 & 35.69 & 14.02\% & 21.78 & 34.92 & 13.40\% & 21.96 & 35.37 & 13.65\% & 21.65 & 34.16 & 13.23\% \\
\multicolumn{2}{l}{STGODE} & 22.61 & 38.34 & 14.28\% & 22.09 & 37.04 & 14.08\% & 22.21 & 37.21 & 15.17\% & 21.85 & 36.86 & 14.65\% \\
\rowcolor[rgb]{ .929, .929, .929} \multicolumn{2}{l}{PASTN} & \textbf{20.43} & \textbf{34.36} & \textbf{13.12\%} & \textbf{19.95} & \textbf{33.47} & \textbf{12.98\%} & \textbf{20.48} & \textbf{33.42} & \textbf{13.06\%} & \textbf{20.06} & \textbf{33.21} & \textbf{12.65\%} \\
\bottomrule
\end{tabular}
\end{adjustbox}
\caption{Performance comparison on short-term vs. long-term holiday traffic patterns. Single Day: the specific holiday date. Entire Period: the full multi-day holiday break.}
\label{tab:holiday}
\end{table}

\noindent\textbf{Performance of Long-Term Predictions.} We evaluate the long-term prediction performance of our model over different horizons. Results in Table~\ref{tab:long-term perform} reveal that PASTN excels in both short-term and long-term traffic predictions, despite the increased challenges associated with extended forecasting periods. The capability of PASTN to provide early warnings allows managers sufficient lead time to proactively mitigate congestion, optimize traffic flow, and ensure road safety.

\begin{table}[htbp]
\setlength{\tabcolsep}{5pt} 
\renewcommand{\arraystretch}{0.85} 
\centering

\begin{adjustbox}{center}
\scriptsize
\begin{tabular}{clccccccccc}
\toprule
\multirow{2}{*}{Model} & Horizon &
 \multicolumn{3}{l}{15min} & \multicolumn{3}{l}{30min} & \multicolumn{3}{l}{1h} \\
  \cmidrule(lr){3-5}\cmidrule(lr){6-8}\cmidrule(lr){9-11}
 & Metrics & MAE & RMSE & MAPE & MAE & RMSE & MAPE & MAE & RMSE & MAPE\\
 \midrule
 \multicolumn{2}{c}{DCRNN}	&18.41	&\underline{29.23}	&10.94\%	&23.16	&36.15	&14.14\%	&30.26	&46.85	&19.68\% \\
 \multicolumn{2}{c}{AGCRN} &\underline{17.27}	&29.70	&10.78\%	&\underline{20.38}	&34.82	&\underline{12.70\%}	&\underline{24.59}	&42.59	&\underline{16.03\%}\\
\multicolumn{2}{c}{STGCN}	&19.86	&34.10	&12.40\%	&22.75	&38.91	&14.11\%	&26.70	&45.78	&17.00\%\\
\multicolumn{2}{c}{GWNET} &17.28	&27.68	&\textbf{10.18\%}	&21.31	&\underline{33.70}	&13.02\%	&26.99	&\underline{42.51}	&17.64\%\\
\multicolumn{2}{c}{STGODE}	&18.10	&30.02	&11.18\%	&21.71	&36.46	&13.64\%	&26.45	&45.09	&17.60\% \\
\rowcolor[rgb]{ .929,  .929,  .929}\multicolumn{2}{c}{PASTN}	&\textbf{16.83}	&\textbf{27.19}	&\underline{10.73\%}	&\textbf{19.88}	&\textbf{32.29}	&\textbf{12.18\%}	&\textbf{24.49}	&\textbf{40.08}	&\textbf{15.86\%}\\
\bottomrule
\end{tabular}
\end{adjustbox}
\caption{Comparison of long-term prediction performance. Underlined values indicate the second-best performance.}
\label{tab:long-term perform}
\end{table}

\section{6 Limitations and Future Directions}
{\textbf{The use of time-series, graph-structured data, and additional information.}}
Current research predominantly relies on graph-structured data and fixed-length historical time-series for traffic forecasting. However, the semantic characteristics of locations, including their functional roles or planned activities, play a crucial role in shaping human mobility patterns~\cite{liang2024exploring}. The generalization experiments in Section 5.3, where models demonstrated improved performance during the pandemic, further highlight the influence of latent environmental semantics on prediction accuracy. Although short-term traffic patterns, typically analyzed through fixed-length time series, are the main focus, long-term periodic trends, such as seasonal and annual cycles, are often overlooked.

Models designed for large-scale traffic forecasting already face significant challenges in generalization and scalability, even when restricted to basic data inputs like graph structures and short-term time-series. These issues are further compounded by computational constraints and frequent memory limitations, significantly reducing the practical feasibility of leveraging richer data sources. Addressing these challenges necessitates future work focused on systematically identifying the factors most critical to improving predictive accuracy and determining the optimal spatial and temporal resolutions for effective and scalable modeling. Moreover, representing and utilizing semantic information effectively remains an open problem. Multimodal signals, including textual descriptions of planned activities and visual representations of road layouts, demand novel approaches to ensure both predictive performance and computational efficiency.

\noindent{\textbf{The development of simple yet effective approaches.}} Analysis of the results presented in Table~\ref{tab:overall performance} and Table~\ref{tab:efficiency} reveals that the increasing complexity of recent models has led to significant inefficiencies and limited scalability in large-scale sensor networks. A shift towards methods that prioritize computational simplicity without compromising predictive performance is crucial for overcoming these limitations and enabling broader applicability in real-world resource-intensive scenarios.

One potential solution lies in knowledge distillation, which facilitates the compression of large, complex models (teacher models) into smaller, efficient ones (student models) while preserving predictive accuracy~\cite{hinton2015distilling}. By alleviating computational overhead, this approach may enhance scalability and ensure the feasibility of deploying predictive models across extensive networks.
Another promising research direction may focus on the creation of spatiotemporal foundation models, inspired by breakthroughs in other domains, such as ChatGPT in natural language processing~\cite{wu2023gpt} and Segment Anything in computer vision~\cite{kirillov2023segment}. By pre-training on large and diverse spatiotemporal datasets, these models can generalize across heterogeneous traffic scenarios, minimizing the reliance on task-specific re-training. Additionally, their ability to process multimodal data—ranging from textual descriptions to visual representations—offers a unified framework for capturing intricate spatiotemporal patterns, advancing both scalability and efficiency.

\noindent{\textbf{Advancing practical application frameworks.}} Despite the advancements in optimizing models for multi-sensor time-series data under mathematically defined frameworks, limited efforts have been made toward constructing application-oriented frameworks for real-world deployment. Addressing this gap requires an exploration of how these models can align with travel services and traffic management systems to enhance practical usability. Future research could address how models might interact with fine-grained, real-time route planning to provide adaptive navigation solutions or inform dynamic traffic management strategies. Another important avenue may be leveraging real-time traffic data and policy updates to dynamically adjust predictions with agent-based frameworks~\cite{chen2023agentverse}. By enabling agents to simulate and respond to environmental changes, such a framework would enhance the system's adaptability to complex scenarios.

\section{7 Conclusion}
In this paper, we propose a Positional-aware Spatio-Temporal Network (PASTN) addressing critical challenges in large-scale traffic forecasting. This model innovatively incorporates adaptive spatial positional embeddings and an attention-enhanced temporal learning module, enabling it to distinguishably learn unique node characteristics and complex traffic flow patterns across extensive datasets. Extensive benchmarking results show that PASTN surpasses SOTA methods by achieving up to an 18.45\% improvement in RMSE, demonstrating a favorable performance-efficiency trade-off. Further analysis confirms the model's robustness during critical time periods and across key locations, as well as its effectiveness in long-term traffic prediction. PASTN provides a novel and effective solution for managing large-scale transportation systems. Future research could benefit from further exploration of the effective utilization and potential applications of additional multimodal data, the development of simpler yet effective approaches, and the advancement of practical application frameworks to expand the capabilities and impact of traffic forecasting systems.
 
\section{AUTHOR CONTRIBUTIONS}
The authors confirm contribution to the paper as follows: study conception, methodology, experiments and manuscript preparation: {Runfei Chen; result analysis: Shuyang Jiang. Both authors reviewed the results and approved the final version of the manuscript.

\section{Declaration of Conflicting Interests}
The authors declared no potential conflicts of interest with respect to the research, authorship, and/or publication of this article.

\section{Funding}
The authors disclosed receipt of the following financial support for the research, authorship, and/or publication of this article: This research was supported by National Natural Science Foundation of China (Grant No.42171452).


\newpage
\bibliographystyle{trb}
\bibliography{trb_template}
\end{document}